%% file: pfaam.tex
\pdfoutput=1
\documentclass[twocolumn]{article}
\usepackage{arxiv}

\usepackage[utf8]{inputenc} 
\usepackage[T1]{fontenc}    
\usepackage{hyperref}       
\usepackage{url}            
\usepackage{booktabs}       
\usepackage{amsfonts}       
\usepackage{nicefrac}       
\usepackage{microtype}      
\usepackage{cleveref}       
\usepackage{lipsum}         
\usepackage{graphicx}
\usepackage[square,sort,comma,numbers]{natbib}
\usepackage{doi}

\creflabelformat{*}{\textcolor{darkred}{#1}}
\title{Parameter-Free Average Attention Improves Convolutional Neural Network Performance (Almost) Free of Charge}

\author{{Nils Koerber}\\
	Center for Artificial Intelligence in Public Health Research \\
	Robert Koch Institute\\
	Berlin\\
	Germany\\
	\texttt{KoerberN@rki.de} \\
}

\date{}
\renewcommand{\headeright}{}
\renewcommand{\undertitle}{}

\begin{document}

\twocolumn[
  \begin{@twocolumnfalse}
\maketitle

\input{abstr}
\vspace*{0.5cm}
  \end{@twocolumnfalse}
]

\input{introduction}
\input{background}
\input{model}

\input{experiments}
\input{conclusion}

\bibliographystyle{unsrt}  
\bibliography{ref}

\end{document}

%% file: abstr.tex
\begin{abstract}

Visual perception is driven by the focus on relevant aspects in the surrounding world. To transfer this observation to the digital information processing of computers, attention mechanisms have been introduced to highlight salient image regions. Here, we introduce a parameter-free attention mechanism called PfAAM, that is a simple yet effective module. It can be plugged into various convolutional neural network architectures with a little computational overhead and without affecting model size. PfAAM was tested on multiple architectures for classification and segmentic segmentation leading to improved model performance for all tested cases. This demonstrates its wide applicability as a general easy-to-use module for computer vision tasks. The implementation of PfAAM can be found on \url{https://github.com/nkoerb/pfaam}.
	
\end{abstract}

%% file: introduction.tex
\section{Introduction}

Convolutional neural networks have demonstrated an impressive ability to solve a broad range of computer vision tasks \cite{krizhevsky2012imagenet, long2015fully, redmon2016you, he2017mask}. Typically, a convolutional neural network is built modular and the local receptive field is increasing step-wise with network depth. According to that architecture, the network captures hierarchical patterns based on input image representations within the network. Increasing the representational power of neural networks is of ongoing research interest to emphasize the most important features for a given task. Previous work has shown improvements based on adaptations regarding the inner connectivity \cite{huang2017densely, chollet2017xception} or in utilizing an attention mechanism that globally highlight relevant features \cite{hu2018squeeze, woo2018cbam, wang2020eca}. \par
However, existing attention mechanisms rely on trainable parameters, only regard spatial or channel-wise attention, or introduce additional tunable hyperparameters. Here, we introduce Parameter-free Average Attention Module (PfAAM) which improves performance solely by basic mathematical operations and is based on averaging input feature maps. PfAAM can be introduced into network architectures of arbitrary form and do not add trainable parameters or non-trainable hyperparameters and therefore do not change the overall size or complexity of the network. Furthermore, we show that the network performance of different architectures is enhanced by using PfAAM for both classification and semantic segmentation tasks. While most previous work has focused on hand-crafted modules with additional parameters, we present PfAAM as a lightweight plug-and-play module that is compatible with most neural network architectures, enhancing their performance and can be used for various computer vision tasks.

%% file: background.tex
\section{Related Work}

\begin{figure*}[h]
\includegraphics[width=\textwidth]{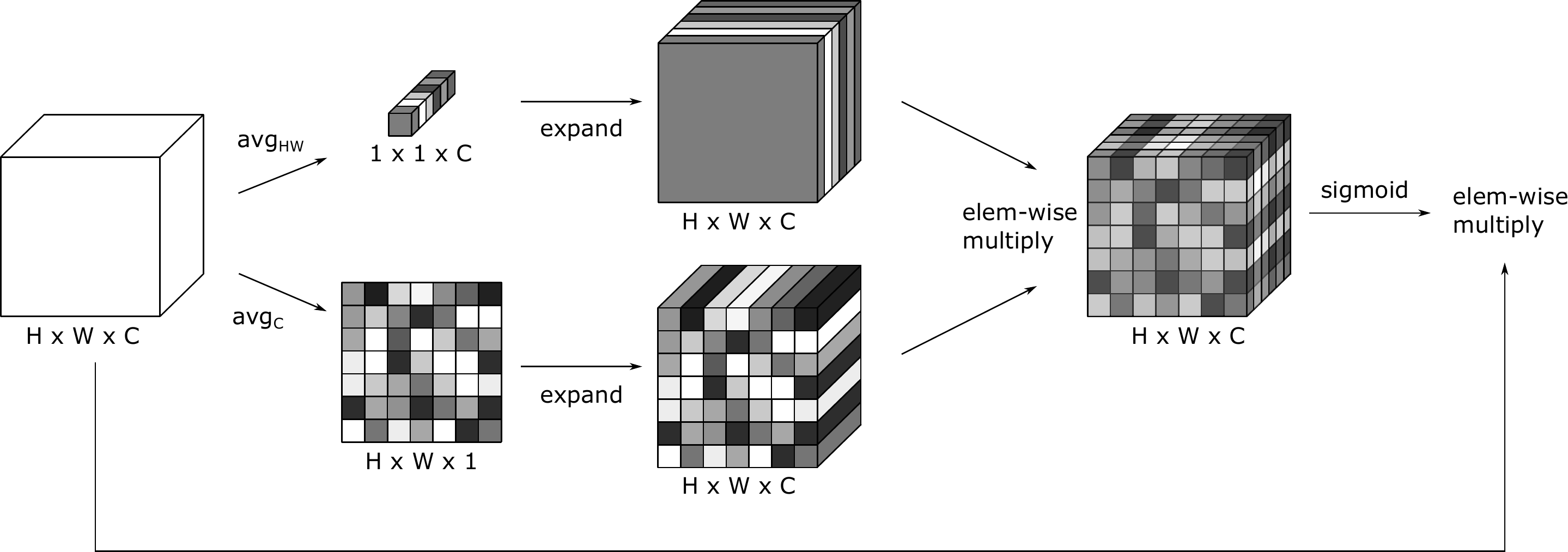}
\caption{Parameter-free Average Attention Module (PfAAM).}
\label{fig:pfaam}
\end{figure*}

In this section, we briefly review related model architectures and attention mechanisms.

\paragraph{Network Architecture.}

With the release of AlexNet \cite{krizhevsky2012imagenet} in 2012 for ImageNet LSVRC-2012 competition \cite{russakovsky2015imagenet}, outperforming all other submissions by a large margin, convolutional neural networks became state-of-the-art for computer vision tasks, having only recently been challenged by vision transformer models \cite{dosovitskiy2020image}, which, however, require expensive pre-training on huge amounts of data. Several improvements to the neural network architecture have been proposed to improve their discriminative abilities. The application of deeper architectures \cite{simonyan2014very, szegedy2015going, wang2017residual}, wider networks \cite{zagoruyko2016wide}, increased connectivity \cite{huang2017densely}, grouped convolutions \cite{xie2017aggregated}, depthwise convolution \cite{xie2017aggregated} or reduced computational requirements \cite{howard2017mobilenets, sandler2018mobilenetv2} have lead to a plethora of potential architectures. Attempts have been made to automatically search for the best network architecture reducing the manual design of neural network architectures \cite{tan2019efficientnet, real2017large}. However, building blocks that boost network performance at low computational costs and can be incorporated in any network architecture without the need for manual adjustment or hyperparameter search remain of high value and motivated the design of PfAAM. \par

\paragraph{Attention Modules.} Human perception is highly selective and filters information based on its relevance for decision making. According to this, so called attention mechnisms have been proposed for computer vision models. See \cite{guo2022attention} for a comprehensive overview of attention mechanisms. Generally, attention mechanisms can be split into methods related to channel attention \cite{hu2018squeeze, wang2020eca} focusing on 'what is important' in the image and spatial attention \cite{oktay2018attention} highlighting 'where is the important information' in the image or a combination of both \cite{woo2018cbam,yang2021simam}. However, most of these attention modules are implemented by adding learnable model parameters \cite{hu2018squeeze, wang2020eca, oktay2018attention, woo2018cbam} during training increasing computational cost and model size, relate only on spatial or channel attention, or depend on tunable hyperparameters \cite{yang2021simam}. As an extension to existing modules, PfAAM captures channel and spatial attention without adding parameters or hyperparameters and is simple by design, promoting a self-reinforcing effect by averaging activations.

%% file: model.tex
\section{Parameter-free Average Attention Module}


The general structure and computation of PfAAM is shown in figure \ref{fig:pfaam}.
Consider a feature map $F \in  \mathbb{R} ^{H \times W \times C} $ as intermediate input, the PfAAM separates the input in a spatial attention part $A_{sp} \in  \mathbb{R} ^{H \times W \times 1}$ by averaging the input along its channels and a channel attention part $A_{ch} \in  \mathbb{R} ^{1 \times 1 \times C}$ by calculating the average of each channel. The resulting attention maps are then expanded along their reduced dimensions and recombined to depict the attention to the most important parts of the feature input map. The final recombined attention map uses a sigmoidal gating mechanism to enhance the representational power of the input.

The overall process can be summarized as follows:

\begin{equation}
F' = \sigma ( A_{sp} \otimes A_{ch} ) \otimes F,
\end{equation}

with $\otimes$ denoting the element-wise multiplication, $\sigma$ the sigmoid function $\sigma(x) = \frac{1}{1+e^{-x}}$ and $F'$ the output of PfAAM. By element-wise multiplication of $A_{sp}$ and $A_{ch}$ the values are broadcasted (copied) along the axes by which they were reduced during averaging to regain their input sizes.

In contrast to attention modules that emphasize features by learned parameters PfAAM is parameter-free and is solely highlighting features via averaging spatially and along channels.

\subsection{Spatial Attention Component}

To emphasize the spatial attention in a feature map we produce a spatial attention map. This is performed by averaging the spatial features along their channels. As a result, the attention is focused on parts in the feature map where a feature is detected.
The average of each spatial element $x_{H \times W} \in \mathbb{R} ^{C}$ can be calculated as follows:
\begin{equation}
A_{sp}(x_{H \times W}) = \frac{1}{C}\sum\limits_{i=1}^C x_{H \times W}(i).
\end{equation}

By averaging along the channels, the dimension is reduced and produces a spatial map where each element represents the average across channels. As a consequence the spatial areas with high activations are emphasized while areas with low activations are suppressed, thus highlighting positions with detected features.

\subsection{Channel Attention Component}
In accordance to spatial attention the channel attention is calculated by averaging along the spatial dimensions of the feature map. Formally, for each channel $y_C \in \mathbb{R} ^{H \times W}$ the average along its spatial dimensions can be calculated as:
\begin{equation}
A_{ch}(y_C) = \frac{1}{H \times W}\sum\limits_{i=1}^H \sum\limits_{j=1}^W y_C(i,j).
\end{equation}

By averaging along the spatial dimensions, channels are emphasized in which a feature is detected and reducing the influence of channels with low activations for their corresponding features.

\subsection{Model Integration}
The three-dimensional input to PfAAM is processed into a matrix with the same dimensions, which can be used as an element-wise multiplier to amplify the activations within the input.  Because of its simplicity, the PfAAM block can be easily integrated into different network architectures and positions, allowing it to be used as a general building block for convolutional neural networks. In the following section, the position of PfAAM within the residual blocks and different pooling operations were analyzed, showing that averaging is slightly preferred over maximization. Finally, PfAAM was successfully tested in different network architectures for classification and segmentation, showing an increase in performance.

%% file: experiments.tex
\section{Experiments}

In this section we tested the optimal PfAAM setup and network integration in an ablation study and performed experiments for classification and semantic segmentation with different network architectures.

\subsection{Ablation Study}

To maximize the effect of the PfAAM block we tested different implementation options. First, we tested averaging versus maximizing as channel and spatial pooling operations within PfAAM to analyze their effect on the overall performance. Max pooling enhance the effect of individual strong activation whereas averaging increases areas with overall strong activation. The performance of a baseline ResNet-164 \cite{he2016deep} with PfAAM blocks added to each residual block of the network were compared based on classification error. Furthermore, we tested the influence of an additional Batch Normalization \cite{ioffe2015batch} before each PfAAM. The accuracy of the network was tested using the CIFAR-10 dataset \cite{krizhevsky2009learning}, that consists of 50k training and 10k test images with a size of 32 x 32 RGB pixels belonging to 10 different classes. See table \ref{tab:ablation} for classification errors of the PfAAM implementations. In total, there is no large difference in the resulting classification error and all PfAAM implementations improve the performance compared to the baseline model, from which we concluded that each of the implementations performs reasonably well. Averaging without Batch Normalization showed the best performance overall, which is why we continued further experiments with this configuration, unless stated otherwise.

\begin{table}
\begin{center}
\caption{Comparison of different PfAAM implementations CIFAR-10 using averaging or maximizing and an additional Batch Normalization (BN). The lowest classification error is shown in bold.}
\begin{tabular}{ l | c }
\hline
 &   error (\%) \\
\hline
ResNet-164 \cite{he2016deep} & 5.46 \\
ResNet-164+PfAAM(max)  & 4.79  \\
ResNet-164+PfAAM(avg) & \textbf{4.76} \\
ResNet-164+BN+PfAAM(max)  & 4.94  \\
ResNet-164+BN+PfAAM(avg)  & 4.86 \\
\hline
\end{tabular}
\label{tab:ablation}
\end{center}
\end{table}

\subsection{Experiments}

To analyze the effect of PfAAM on neural network performance we used baseline architectures for classification and semantic segmentation and compared the performance of the regular architecture to the same architecture but with additional PfAAM blocks incorporated.

\subsubsection{Image Classification}

To investigate the effect of PfAAM in a classification task, we conducted experiments using CIFAR-10 and CIFAR-100 as benchmark. Both data sets have the same size, but are divided into 10 and 100 classes, respectively. As model architectures Residual Networks \cite{he2016deep} and Wide Residual Networks \cite{zagoruyko2016wide} with varying depth and width were used to cover basic architectures from shallow to deep and thin to wide. Results in table \ref{tab:exp} show an reduction of the classification error for all tested architectures with integrated PfAAM. For deeper architectures the effect of PfAAM is larger showing a reduction of the error rate of  over 12\% for ResNet-110 and ResNet-164 on CIFAR-10, thus ResNet-110+PfAAM almost matches the performance of the regular ResNet-164 which has 40\% more trainable parameters. For wider but shallower architectures with fewer residual blocks, the effect of PfAAM is smaller (1.4\% reduction for WRN-16-8 on CIFAR-10), suggesting that the effect scales with the number of PfAAM units per network. Since PfAAM does not introduce additional learnable parameters, it generally increases the performance of the network in image classification by improving the utilization of the existing parameters.

\begin{table}
\begin{center}
\caption{Classification error (\%) on CIFAR-10 and CIFAR-100. The lowest error per model architecture and data set is shown in bold.}
\begin{tabular}{ l | c | c | c }

\hline
\multicolumn{4}{c}{CIFAR-10}\\
\hline
 &  \# params & original  & +PfAAM\\
\hline
ResNet-110 \cite{he2016deep} & 1.2M & 6.37 & \textbf{5.57} \\
ResNet-164 \cite{he2016deep} & 1.7M & 5.46 &\textbf{4.76} \\
WRN-28-2 \cite{zagoruyko2016wide} & 1.5M & 5.73 & \textbf{5.29} \\
WRN-16-8 \cite{zagoruyko2016wide} & 11M & 4.27 & \textbf{4.21} \\
\hline
\multicolumn{4}{c}{CIFAR-100}\\
\hline
 &  \# params & original  & +PfAAM\\
\hline
ResNet-110 \cite{he2016deep} & 1.2M & 26.88 & \textbf{24.22} \\
ResNet-164 \cite{he2016deep} & 1.7M & 24.33 & \textbf{23.05} \\
WRN-28-2 \cite{zagoruyko2016wide} & 1.5M & 26.69 & \textbf{25.38} \\
WRN-16-8 \cite{zagoruyko2016wide} & 11M & 20.43 & \textbf{20.33} \\
\hline
\end{tabular}
\label{tab:exp}
\end{center}
\end{table}

\subsubsection{Semantic Segmentation}

To test PfAAM for semantic segmentation, we used the PASCAL VOC 2012 segmentation dataset \cite{everingham2010pascal} consisting of 1464 training and 1449 validation images of 20 categories and an additional background class. Following previous work \cite{zhao2017pyramid,chen2014semantic, long2015fully}, we used the extended dataset with annotations from \cite{hariharan2011semantic} resulting in 10582 training images. We trained a U-Net \cite{ronneberger2015u} and a Feature Pyramid Network (FPN) \cite{lin2017feature} on the training images and compared the results to the same architectures with added PfAAM. Each model used a ResNet-50 \cite{he2016deep} as encoder-backbone, which was pre-trained on the ImageNet dataset \cite{russakovsky2015imagenet}. The results in table \ref{tab:seg} show the mean intersection over union (mIoU) on the validation images. Both models show increased performance when trained with PfAAM increasing the mIoU by 7.7\% for U-Net and 5.3\% for FPN, respectively. The averaged validation mIoU for U-Net with and without PfAAM during training are depicted in figure \ref{tab:seg} showing a clear improvement for the PfAAM-model. Similar to classification, the introduction of the PfAAM in the model architecture improves the performance, underlining its general applicability as neural network building block enhancing model performance. 

\begin{table}
\begin{center}
\caption{Segmentation results (mIoU, \%) on PASCAL VOC 2012 validation set. Best results per model architecture are shown in bold.}
\begin{tabular}{ l | c | c }
\hline
 &   original & +PfAAM\\
\hline
 U-Net  & 55.7 & \textbf{60.3}\\
 FPN   & 56.5 & \textbf{59.7}\\
\hline
\end{tabular}
\label{tab:seg}
\end{center}
\end{table}

\begin{figure}[h]
\begin{center}
\includegraphics[width=0.5\textwidth]{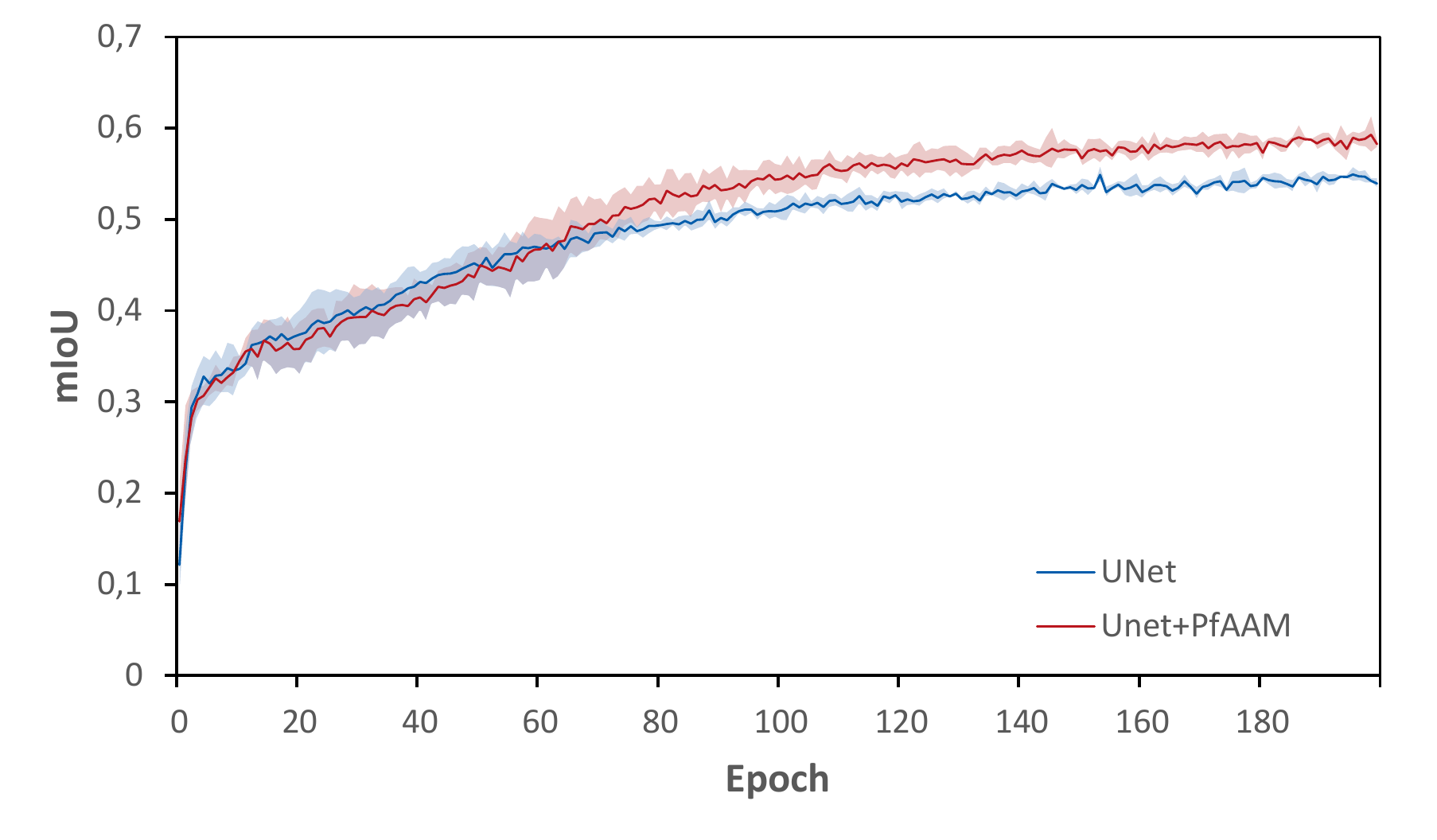}
\caption{Validation mIoU during training for a regular U-Net and the same architecture extended by PfAAM, shown is $mean \pm std$.}
\end{center}
\label{fig:seg}
\end{figure}

\subsubsection{Implementation Details}
For CIFAR training we followed the established standard training procedure used by the original publications \cite{he2016deep,zagoruyko2016wide}. Each 32x32 image or its its horizontally mirrored version was padded by 4 pixels and randomly cropped back to 32x32 pixel. The neural networks were trained for 200 epochs by optimizing the cross-entropy loss using SGD (stochastic gradient decent) with a momentum of 0.9, a weight decay of 0.0005 a mini-batch size of 128 and an initial learning rate of 0.1. The learning rate was step-wise decreased after 60 epochs, 120 epochs and 160 by a factor of 0.2. \par
For semantic segmentation using the PASCAL VOC dataset, the training images were randomly horizontally flipped and scaled by a factor of 0.5 to 2 for each axis, from which random 224x224 patches were cut and fed into the neural network. Optimization was performed using SGD with a momentum of 0.9 and a constant learning rate of 0.0001 for 200 epochs, optimizing the cross-entropy loss function excluding pixels labeled as \textit{void}.\par
Unless stated otherwise, all results are reported as the median over 5 runs.

%% file: conclusion.tex
\section{Conclusion}

In this work, we present a novel attention mechanism PfAAM based on highlighting areas of high activation. When PfAAM is used in different network architectures for classification and semantic segmentation, the performance increases for all tested architectures, while the network size remained unchanged and the computational cost is low. Even though PfAAM does not add additional trainable parameters to the network and does not rely on other theoretical considerations, its positive effect is surprisingly robust, suggesting that it leads to a self-focusing effect on relevant features. In summary, PfAAM provides a simple novel building block that might be considered for future neural network design in computer vision tasks.